%% file: acl2017.tex
\documentclass[11pt,a4paper]{article}
\usepackage[hyperref]{acl2017}
\usepackage{times}
\usepackage{latexsym}

\usepackage{url}

\usepackage{graphicx}
\usepackage{booktabs}

\usepackage{listings}
\usepackage{arabtex}
\usepackage{utf8}
\usepackage[normalem]{ulem}
\setcode{utf8}
\usepackage{comment}

\usepackage{amsmath,amssymb}

\aclfinalcopy 

\usepackage{color}
\newcommand\alert[1]{{\textcolor{red}{#1}}}

\title{Challenging Language-Dependent Segmentation for Arabic: \\An Application to Machine Translation and Part-of-Speech Tagging}

\author{Hassan Sajjad ~~ Fahim Dalvi ~~ Nadir Durrani ~~ Ahmed Abdelali\\ \textbf{Yonatan Belinkov}$^*$~~ \textbf{Stephan Vogel} \\\\
Qatar Computing Research Institute -- HBKU, Doha, Qatar  \\
{\tt \{hsajjad, faimaduddin, ndurrani, aabdelali, svogel\}@qf.org.qa}\\\\
$^*$MIT Computer Science and Artificial Intelligence Laboratory, Cambridge, MA 02139, USA \\
{\tt belinkov@mit.edu}
}

\date{}

\begin{document}
\maketitle
\begin{abstractX}

Word segmentation plays a pivotal role in improving any Arabic NLP application. Therefore, a lot of research has been spent in improving its accuracy. Off-the-shelf tools, however, are: i) complicated to use and ii) domain/dialect dependent. We explore three language-independent alternatives to morphological segmentation using: i) data-driven sub-word units, ii) characters as a unit of learning, and iii) word embeddings learned using a character CNN (Convolution Neural Network). On the tasks of Machine Translation and POS tagging, we found these methods to achieve close to, and occasionally surpass state-of-the-art performance. In our analysis, we show that a neural machine translation system is sensitive to the ratio of source and target tokens, and a ratio close to 1 or greater, gives optimal performance.

\end{abstractX}

\input{introduction}
\input{segmentation}

\input{experiments}
\input{pos}


\section{Conclusion}
\label{sec:conclusion}

We explored several alternatives to language-dependent segmentation of Arabic and evaluated them on the tasks of machine translation and POS tagging. On the machine translation task, {\tt BPE} segmentation produced the best results and even outperformed the state-of-the-art morphological segmentation in the Arabic-to-English direction. On the POS tagging task, character-based models got closest to using the state-of-the-art segmentation. Our results showed that data-driven 
segmentation schemes can serve as an alternative to heavily engineered
language-dependent tools and achieve very competitive results.  In our analysis we showed that NMT performs better when the source to target token ratio is close to one or greater. 

\section*{Acknowledgments}

We would like to thank the three anonymous reviewers for their useful suggestions. 
This research was carried out in collaboration between the HBKU Qatar Computing Research Institute (QCRI) and the MIT Computer Science and Artificial Intelligence Laboratory (CSAIL).

\bibliography{eacl2017,semat2016}

\bibliographystyle{acl_natbib}


\end{document}

%% file: introduction.tex
\section{Introduction}


Arabic word segmentation has shown to significantly improve 
output quality in NLP tasks such as machine translation \cite{habash:naacl06,almahairi:arxiv},
part-of-speech 
tagging \cite{Diab:2004:ATA:1613984.1614022,Habash:2005:ATP:1219840.1219911}, and information retrieval \cite{aljlayl:2002}. A considerable amount of research has therefore
been spent on Arabic morphological segmentation in the past two decades, ranging from rule-based analyzers  \cite{
beesley1996arabic} 
to state-of-the-art statistical segmenters \cite{
pasha2014madamira,abdelali-EtAl:2016:N16-3,khalifayamama}.  
%
Morphological segmentation splits words into morphemes.  
For example, 
`\mbox{`\textit{wktAbnA}''}
``\<وكتابنا>'' (gloss: and our book) 
is decomposed into its stem and affixes as: 
\mbox{``\textit{w+ ktAb +nA}''} ``\<و+ كتاب +نا>''.

Despite the gains obtained 
from using 
morphological segmentation, there are several caveats to 
using these tools. Firstly, 
they make
the training pipeline 
cumbersome, as they 
come with complicated pre-processing 
(and additional post-processing in the case of English-to-Arabic translation~\cite{Kholy:Habash:2012}). More importantly, 
these tools are dialect- and domain-specific. A segmenter trained for 
modern standard Arabic (MSA) 
performs significantly worse on dialectal Arabic \cite{Habash13morph}, or when it 
is applied 
to a new domain. 


In this work, we explore whether we can avoid the
\emph{language-dependent} pre/post-processing components and learn segmentation directly from the training data 
being used for a given task. 
We investigate data-driven alternatives to morphological 
segmentation using i) 
unsupervised sub-word units obtained using 
byte-pair encoding 
\cite{sennrich-haddow-birch:2016:P16-12}, 
ii) purely character-based segmentation \cite{lingTDB15}, 
and iii) a convolutional neural network 
over characters~\cite{KimAAAI1612489}. 

We evaluate these techniques on the tasks of machine translation (MT) and part-of-speech (POS) tagging and compare them against 
morphological segmenters MADAMIRA \cite{pasha2014madamira} and Farasa \cite{abdelali-EtAl:2016:N16-3}. On the MT task, 
byte-pair encoding ({\tt BPE})
performs the best among the three methods, achieving very similar performance to morphological segmentation in the Arabic-to-English direction and slightly worse in the other direction. 
Character-based methods, in comparison, perform better on the task of POS tagging, 
reaching 
an accuracy of $95.9\%$, only $1.3\%$ worse 
than 
morphological segmentation. 
We also analyze the effect of segmentation granularity of Arabic on the quality of MT. We observed that a neural MT (NMT) system is sensitive to source/target token ratio and performs best when this ratio is close to or greater \mbox{than 1}.

%% file: segmentation.tex
\section{Segmentation Approaches}
\label{sec:segmentation}

We experimented with three data-driven segmentation schemes: 
i) morphological segmentation, ii) sub-word segmentation based on \texttt{BPE}, 
and iii) two variants of character-based segmentation.
%
We first map each source word to its corresponding segments (depending on the segmentation scheme), 
 embed all segments of a word in vector space and feed them one-by-one to an encoder-decoder 
 model. See Figure \ref{fig:segmentation} for illustration. 

\subsection{Morphological Segmentation}

There is a vast amount of work on statistical segmentation for Arabic. Here we use the state-of-the-art Arabic segmenter MADAMIRA and Farasa as our baselines.
MADAMIRA involves a morphological analyzer that generates a list of possible word-level analyses (independent of context). The analyses are provided with the original text to a {\tt Feature Modeling} component that applies 
an
SVM 
and 
a language model 
to make predictions, which are scored by an {\tt Analysis Ranking} component. 
Farasa on the other hand is a light weight 
segmenter, which ignores 
context and instead uses a variety of features and lexicons for segmentation. 


\subsection{Data Driven Sub-word Units}

%
A number of data-driven 
approaches have been proposed 
that learn
to segment words into smaller units from data \cite{demberg:07,virpioja:13} and 
shown to improve phrase-based MT \cite{Fishel10linguisticallymotivated,Stallard:2012:UMR}.
Recently, with the advent of neural MT, a few sub-word-based techniques have been proposed that segment words into 
smaller
units to tackle the limited vocabulary 
and unknown word problems \cite{sennrich-haddow-birch:2016:P16-12,google-nmt:2016}. 

In this work, we explore \emph{Byte-Pair Encoding} ({\tt BPE}), a data compression algorithm \cite{Gage:1994:NAD:177910.177914} as an alternative to morphological segmentation of Arabic.
%
\texttt{BPE} splits words into symbols (a sequence of characters) and then iteratively replaces the most frequent symbols with their merged variants. In essence, frequent character n-gram sequences will be merged to form one symbol. The number of merge operations is controlled by a hyper-parameter {\tt OP} which directly 
%
affects the granularity of segmentation: 
a high value 
of {\tt OP} 
means coarse segmentation and a low value means fine-grained segmentation.



\begin{figure}[t] 
\centering
\includegraphics[width=0.95\linewidth]{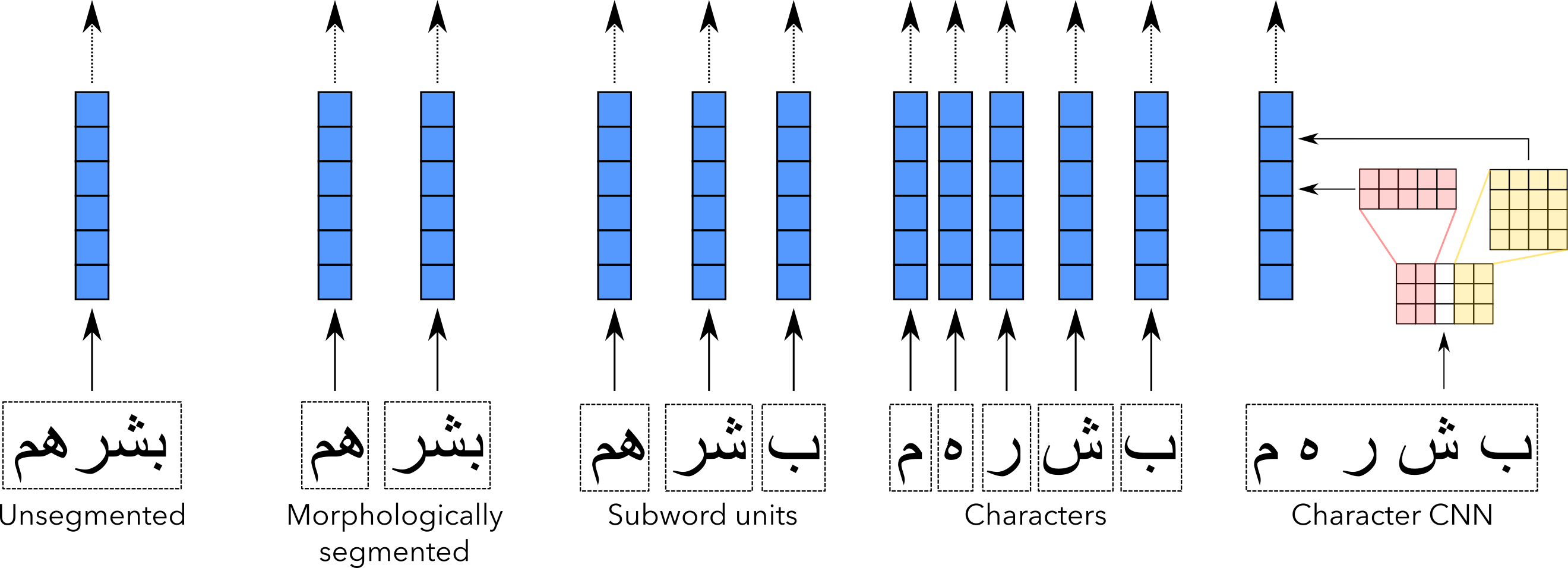}
\caption{Segmentation approaches 
for the word ``b\$rhm'' ``\<بشرهم>''; the blue vectors indicate the embedding(s) used before the encoding layer. 
}
\label{fig:segmentation}
\end{figure}

\subsection{Character-level Encoding}

Character-based models have been found to be effective in translating closely related language pairs \cite{durrani-EtAl:2010:ACL,Nakov:Tiedemann:2012} and OOV words \cite{durrani-EtAl:2014:EACL}. \newcite{charNMT2016:wang}
used character embeddings to address the OOV word problem. We explored them as an alternative to morphological segmentation. Their advantage is that character embeddings do not require any complicated pre- and post-processing step other than segmenting words into characters. The fully character-level encoder treats the source sentence as a sequence of letters, encoding each letter (including white-space) in the LSTM 
encoder (see Figure \ref{fig:segmentation}). The decoding may follow identical settings. 
We restricted the character-level representation to the Arabic side of the parallel corpus and use words for the English side. 

\paragraph{Character-CNN}

~\newcite{KimAAAI1612489} presented a neural language model that takes character-level input and learns word embeddings using a CNN over characters. The embedding are then provided to the encoder as input. The intuition is that the character-based word embedding should be able to learn the morphological phenomena 
a word inherits. Compared to fully character-level encoding, the encoder gets 
word-level embeddings as in the case of unsegmented words (see Figure \ref{fig:segmentation}). However, the word embedding is intuitively richer than the embedding learned over unsegmented words because of the convolution over characters.
The method was previously shown to help 
neural MT  \cite{belinkov:semat2016,costajussa-fonollosa:2016:P16-2}.
\newcite{belinkov:2017:ACL} also showed character-based representations learned using a CNN 
to be superior, at learning word morphology, than their word-based counter-parts. However, they did not compare these against BPE-based segmentation.
We use character-CNN to aid Arabic word segmentation. 

%% file: experiments.tex
\begin{table*}[!ht]
\centering
\footnotesize
\begin{tabular}{l|rrrr|r||rrrr|r}
\toprule
 & \multicolumn{5}{c||}{Arabic-to-English} & \multicolumn{5}{c}{English-to-Arabic} \\
\# SEG & tst11 & tst12 & tst13 & tst14 & AVG. & tst11 & tst12 & tst13 & tst14 & AVG. \\
\midrule
{\tt UNSEG} & 25.7 & 28.2 & 27.3 & 23.9 & 26.3 & 15.8 & 17.1 & 18.1 & 15.5 & 16.6 \\
\midrule
{\tt MORPH} & 29.2 & 33.0 & 32.9 & 28.3 & 30.9 & 16.5 & 18.8 & 20.4 & 17.2 & {\bf 18.2} \\
{\tt cCNN} & 29.0 & 32.0 & 32.5 & 28.0 & 30.3 & 14.3 & 12.8 & 13.6 & 12.6 & 13.3 \\
{\tt CHAR} & 28.8 & 31.8 & 32.5 & 27.8 & 30.2 & 15.3 & 17.1 & 18.0 & 15.3 & 16.4 \\
{\tt BPE}  & 29.7 & 32.5 & 33.6 & 28.4 & {\bf 31.1} & 17.5 & 18.0 & 20.0 & 16.6 & 18.0 \\ 
\bottomrule
\end{tabular}
\caption{\label{tab:results-ar-en} Results of comparing several segmentation strategies. 
}
\end{table*}

\section{Experiments}
\label{sec:experiments}
In the following, we describe the data and system settings and later present the results of machine translation and POS tagging. 
\subsection{Settings}

\paragraph{Data} The MT systems were trained on 1.2 Million sentences, a concatenation of TED corpus \cite{cettoloEtAl:EAMT2012}, 
LDC NEWS data, QED \cite{guzman-sajjad-etal:iwslt13} 
and an MML-filtered \cite{Axelrod_2011_emnlp} UN corpus.\footnote{We used 3.75\% as reported to be optimal filtering threshold in \cite{durrani-etal:2016:IWSLT}.}
%
We used 
dev+test10 
for tuning and 
tst11-14 
for testing. For English-Arabic, outputs were detokenized using MADA detokenizer. Before scoring the output, we normalized them and reference translations using the QCRI normalizer \cite{sajjad-etal:iwslt13}.
\paragraph{POS tagging} We used 
parts 2-3 (v3.1-2) 
of the Arabic Treebank
\cite{
maamouri2011atb3}. 
The data consists of 18268 sentences (483,909 words).
We 
used 80\% for training, 5\% for development and the remaining 
for test.


\paragraph{Segmentation}  
MADAMIRA and Farasa normalize the data before segmentation. In order to have consistent data, we normalize it for all segmentation approaches.  For {\tt BPE}, we tuned the value of merge operations {\tt OP} and found 30k and 90k to be optimal for Ar-to-En and En-to-Ar respectively.
In case of \textit{no segmentation} ({\tt UNSEG}) and \textit{character-CNN} ({\tt cCNN}), we tokenized the Arabic with the standard Moses tokenizer, which separates punctuation marks.
For \textit{character-level} encoding ({\tt CHAR}), we preserved word boundaries by replacing space with a special symbol and then separated every character with a space. English-side is  tokenized/truecased using Moses scripts.

\paragraph{Neural MT Settings} We used the \emph{seq2seq-attn}~\cite{kim2016} implementation, 
with 2 layers of LSTM in the (bidirectional) encoder and the decoder,
with a size of 500. 
We 
limit the sentence length to 100 for {\tt MORPH}, {\tt UNSEG}, {\tt BPE}, {\tt cCNN}, and 500 for {\tt CHAR} experiments. 
The source and target vocabularies are limited to 50k each. 

\subsection{Machine Translation Results}

Table \ref{tab:results-ar-en} presents 
MT 
results using various segmentation strategies. 
Compared to the {\tt UNSEG} system,
the {\tt MORPH} system\footnote{Farasa performed better in the Ar-to-En experiments and MADAMIRA performed better in the En-to-Ar direction. We used best results as our baselines for comparison and call them {\tt MORPH}.} improved translation quality by 4.6 and 1.6 BLEU points in Ar-to-En and En-to-Ar systems, respectively. 
The results also improved by up to 3 BLEU points for {\tt cCNN} and {\tt CHAR} systems in the Ar-to-En direction. 
However, the performance is lower by at least 0.6 BLEU points compared to the {\tt MORPH} system. 

In the En-to-Ar 
direction, where {\tt cCNN} and {\tt CHAR} are applied on the target side, the performance dropped significantly. In the case of {\tt CHAR}, 
mapping one source word to many target characters makes it harder for NMT to learn a good model. This is in line with our finding on using a lower value of {\tt OP} for {\tt BPE} segmentation 
(see paragraph \textbf{Analyzing the effect of OP}). 
Surprisingly, the {\tt cCNN} system results were inferior to the {\tt UNSEG} system for En-to-Ar. A possible explanation 
is that the decoder's predictions are still done at word level even when using the {\tt cCNN} model (which encodes the target input during training but not the output). In practice, this can lead to generating unknown words. Indeed, in the Ar-to-En case {\tt cCNN} significantly reduces the unknown words in the test sets, while in the En-to-Ar case the number of unknown words remains roughly the same between {\tt UNSEG}  and {\tt cCNN}. 

The {\tt BPE} system outperformed all other systems in the Ar-to-En direction and is lower than {\tt MORPH} by only 0.2 BLEU points in the opposite 
direction. 
This shows that machine translation involving the Arabic language can achieve competitive results with data-driven segmentation. 
This comes with an additional benefit of \emph{language-independent} pre-processing and post-processing pipeline. 
In an attempt to find, whether the gains obtained from data-driven segmentation techniques and morphological segmentation are additive, we applied BPE to morphological segmented data. We saw further improvement of up to 1 BLEU point by using the two segmentations in tandem.

\paragraph{Analyzing the effect of OP:}

\begin{figure} 
\centering
\includegraphics[width=\linewidth]{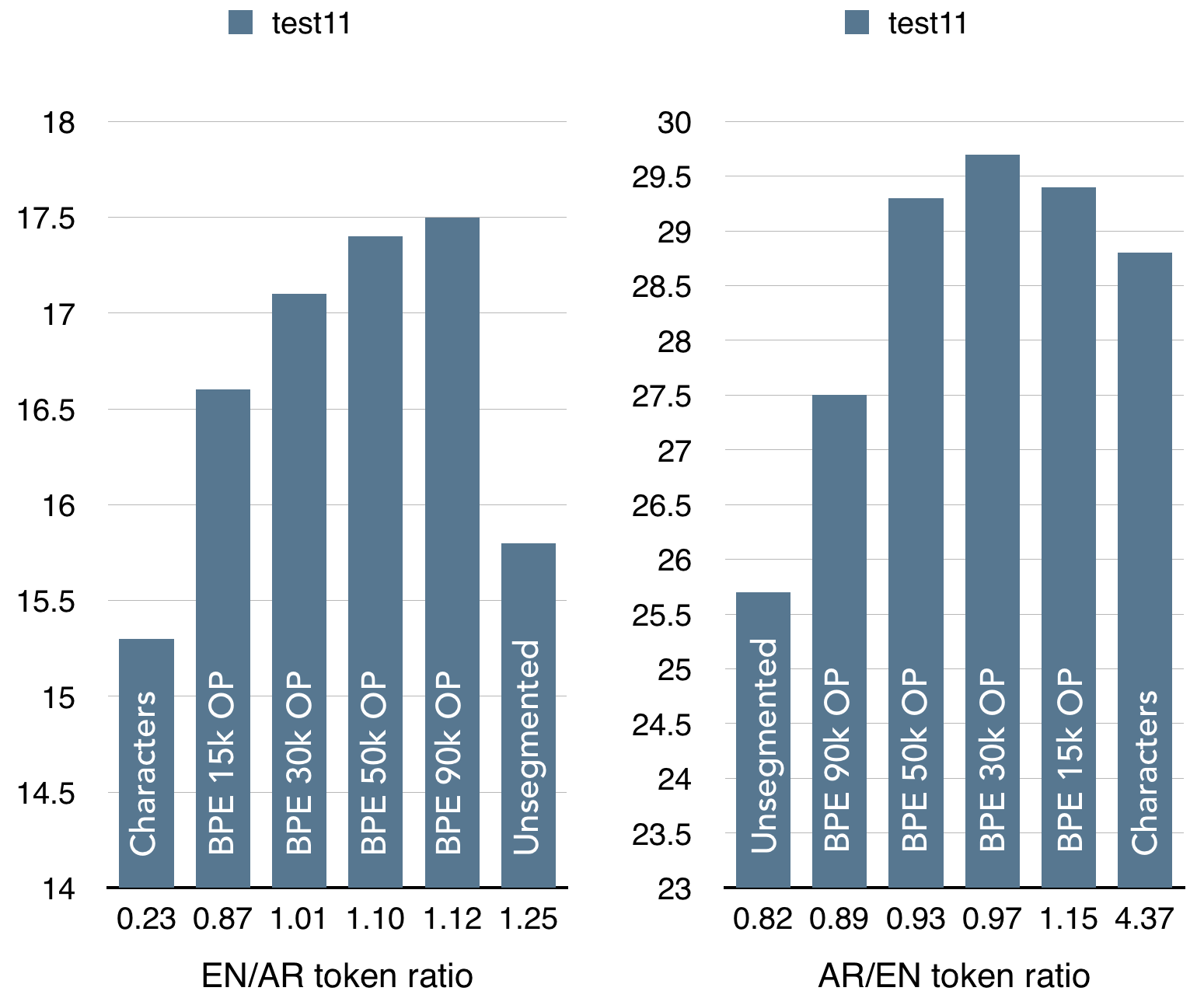}
\caption{Source/Target token ratio with varying {\tt OP} versus BLEU. Character and unsegmented systems can be seen as {\tt BPE} with {\tt OP}=0 and {\tt OP}=$N$.} 
\label{fig:ar-en}
\end{figure}

The unsegmented training data consists of 23M Arabic tokens and 28M English tokens. The parameter {\tt OP} decides the granularity of segmentation: 
a higher value of {\tt OP} means fewer segments. For example, at {\tt OP}=50k, the number of Arabic tokens is greater by 7\% compared to {\tt OP}=90k. We tested four different values of {\tt OP} (15k, 30k, 50k, and 90k). 
Figure \ref{fig:ar-en} summarizes our findings on test-2011 dataset, where x-axis presents the ratio of source to target language tokens and y-axis shows the BLEU score. The boundary values for segmentation are character-level segmentation ({\tt OP}=0) and unsegmented text ({\tt OP}=$N$).\footnote{$N$ is the number of types in the unsegmented corpus.} 
For both language directions, we observed that a source to target token ratio close to 1 and greater works best provided that the boundary conditions (unsegmented Arabic and character-level segmentation) are avoided. In the En-to-Ar 
direction, the system improves for coarse segmentation whereas in the Ar-to-En 
direction, a much finer-grained segmentation of Arabic performed better. 
This is in line with the ratio of tokens generated using the {\tt MORPH} systems (Ar-to-En ratio $=1.02$). 
Generalizing from the perspective of neural MT, the system learns better when total numbers of source and target tokens are close to each other. The system shows better tolerance towards modeling many source words to a few target words compared to the other way around.

\paragraph{Discussion:} Though BPE performed well for machine translation, there are a few reservations that we would like to discuss here. Since the main goal of the algorithm is to compress data and segmentation comes as a by-product, it often produces different segmentations of a root word when occurred in different morphological forms. For example, the words \emph{driven} and \emph{driving} are segmented as \emph{driv en} and \emph{drivi ng} respectively. This adds ambiguity to the data and may result in unexpected translation errors. Another limitation of BPE is that at test time, it may divide the unknown words to semantically different known sub-word units which can result 
in a semantically wrong translation. For example, the word ``\<قطر>'' is unknown to our vocabulary. BPE segmented it into known units which ended up being
translated to \emph{courage}. One possible solution to this problem is; at test time, BPE is applied to those words only which were known to the full vocabulary of the training corpus. In this way, the sub-word units created by BPE for the word are already seen in a similar context during training and the model has learned to translate them correctly. The downside of this method is that it limits BPE's power to segment unknown words to their correct sub-word units and outputs them as \emph{UNK} in translation.

%% file: pos.tex
\subsection{Part of Speech Tagging}
\label{sec:pos}

\begin{figure}[t]
\centering
\includegraphics[width=\linewidth]{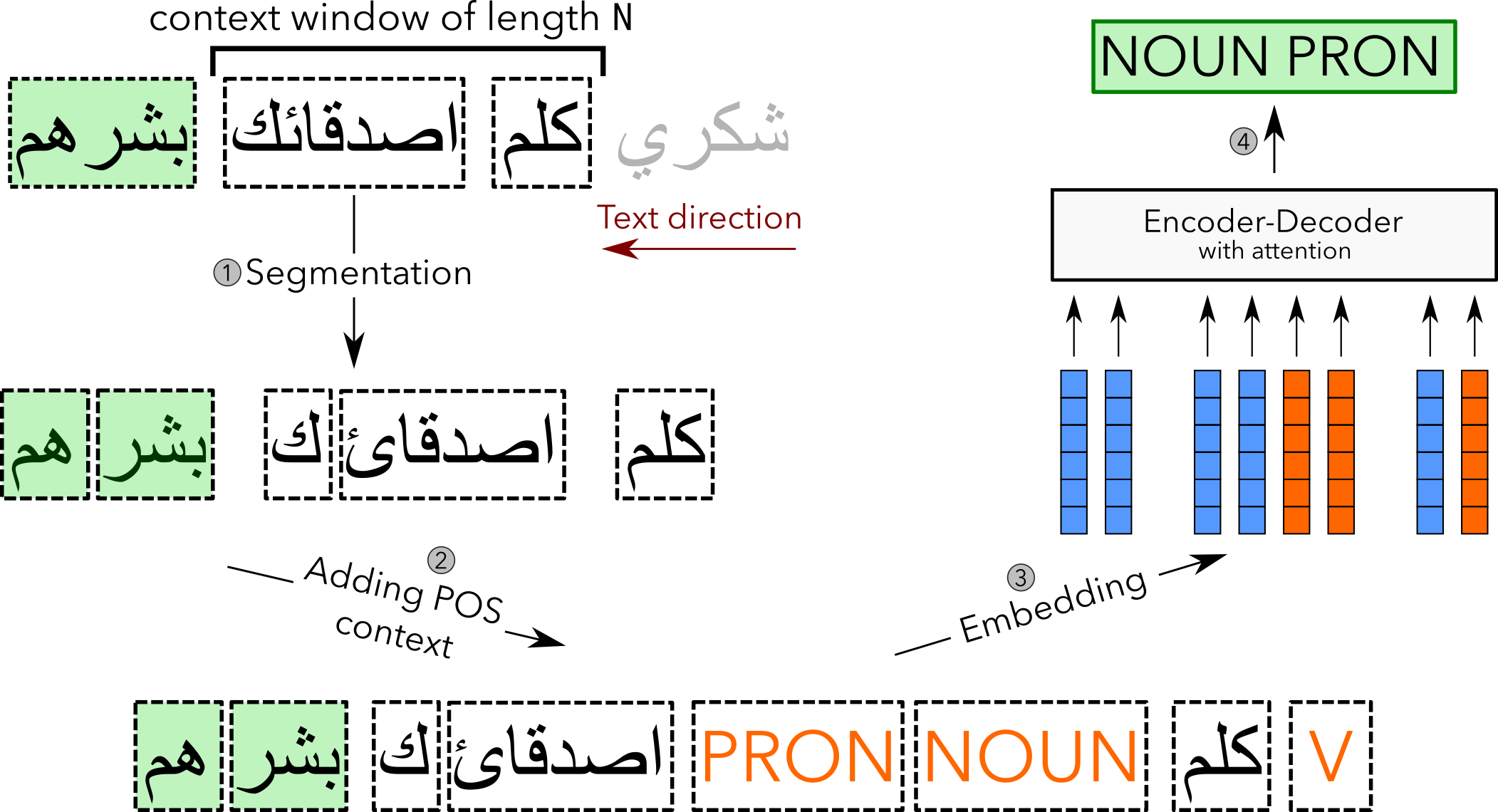}
\caption{Seq-to-Seq POS Tagger: The number of segments and the embeddings depend on the segmentation scheme used (See Figure \ref{fig:segmentation}).}
\label{fig:pos}
\end{figure}

We also experimented with the aforementioned segmentation strategies 
for the task of Arabic POS tagging. Probabilistic taggers like HMM-based \cite{Brants:2000} and sequence learning models like CRF \cite{Lafferty:2001:CRF:645530.655813} consider previous words and/or tags to predict the tag of the current word. We mimic a similar setting but in a sequence-to-sequence learning 
framework.
Figure \ref{fig:pos} describes a step by step procedure to train a neural encoder-decoder tagger. Consider an Arabic phrase 
``klm $>$SdqA\}k b\$rhm'' ``\<كلم أصدقائك بشرهم>'' (gloss: call your friends give them the good news), we want to learn the tag of the word ``\<بشرهم>'' using the context of the previous two words and their tags. First, we segment the phrase using a segmentation approach (step 1) and then add POS tags to context words (step 2). 
The entire sequence with the words and tags is fed to 
the sequence-to-sequence 
framework. The embeddings (for both words and tags) are learned jointly with other parameters in an end-to-end fashion, and 
optimized on the target tag sequence; for example, ``NOUN PRON'' in this case. 

For a given word $w_i$ in a sentence $s=\{w_1,w_2, ..., w_M\}$ and its POS tag $t_i$, We formulate the neural {\tt TAGGER} as follows:

\begin{align*}
	& \texttt{SEGMENTER}(\tau) : \forall w_i \mapsto S_i \\
    & \texttt{TAGGER} : S_{i-2}\ S_{i-1}\ S_i \mapsto t_i
\end{align*}

\noindent where $S_i$ is the segmentation of word $w_i$. In case of {\tt UNSEG} and {\tt cCNN}, $S_i$ would be same as $w_i$. \texttt{SEGMENTER} here is identical to the one described in Figure \ref{fig:segmentation}. \texttt{TAGGER} is a NMT architecture that learns to predict a POS tag of a segmented/unsegmented word given previous two words.\footnote{We also tried using previous words with their POS tags as context but did not see any significant difference in the end result.}

Table \ref{tab:results-pos} summarizes the results. The {\tt MORPH} system performed best with an 
improvement of 5.3\% over {\tt UNSEG}. Among the data-driven methods, {\tt CHAR} model performed 
best and was behind {\tt MORPH} by only 0.3\%. 
Even though {\tt BPE} was inferior 
compared to other 
methods, it was still better than {\tt UNSEG} by 4\%.\footnote{Optimizing the parameter {\tt OP} did not yield any difference in accuracy. We used $10k$ operations.} 


\paragraph{Analysis of POS outputs}

We performed a comparative error analysis of predictions made through {\tt MORPH}, {\tt CHAR} and {\tt BPE} based segmentations. {\tt MORPH} and {\tt CHAR} observed very similar error patterns, with most confusion between \emph{Foreign} and \emph{Particle} tags. In addition to this confusion, {\tt BPE} had relatively scattered errors. It had lower precision in predicting nouns and had confused them with adverbs, foreign words and adjectives. This is expected, since most nouns are out-of-vocabulary terms, and therefore get segmented by {\tt BPE} into smaller, possibly known fragments, which then get confused with other tags. However, since the accuracies are quite close, the overall errors are very few and similar between the various systems. We also analyzed the number of tags that are output by the sequence-to-sequence model using various segmentation schemes. In 99.95\% of the cases, the system learned to output the correct number of tags, regardless of the number of source segments.

\begin{table}[t]
\centering
\footnotesize
\begin{tabular}{r|rr|rrr}
\toprule
SEG & {\tt UNSEG} & {\tt MORPH} & {\tt CHAR} & {\tt cCNN} & {\tt BPE} \\
\midrule
ACC & 90.9 & 96.2 & 95.9 & 95.8 & 94.9 \\
\bottomrule
\end{tabular}
\caption{\label{tab:results-pos} POS tagging with various segmentations  
}
\end{table}

